\documentclass{article}
\usepackage{spconf,amsmath,graphicx}
\usepackage{gensymb, textcomp}
\usepackage{pgfplots}
\usepackage{setspace}
\usepackage{color}
\pgfplotsset{compat=newest}

\definecolor{red}{rgb}{1,0,0}

\title{
Rotation invariant CNN using scattering transform \\for image classification\\
}
\twoauthors
{Rosemberg Rodriguez, Eva Dokladalova}
{ESIEE Paris, University Paris-Est\\
LIGM UMR 8049\\
2 Bd Blaise Pascal, Noisy le Grand, France}
{Petr Dokladal}
{MINES ParisTech - PSL Research University\\
Center for Mathematical Morphology\\
35 rue Saint Honor\'e, Fontainebleau, France}
\begin{document}
%
\maketitle
\begin{abstract}
Deep convolutional neural networks accuracy is heavily impacted by rotations of the input data. In this paper, we propose a convolutional predictor that is invariant to rotations in the input. This architecture is capable of predicting the angular orientation without angle-annotated data. Furthermore, the predictor maps continuously the random rotation of the input to a circular space of the prediction. For this purpose, we use the roto-translation properties existing in the Scattering Transform Networks with a series of 3D Convolutions. We validate the results by training with upright and randomly rotated samples. This allows further applications of this work on fields like automatic re-orientation of randomly oriented datasets.

\end{abstract}

\section{Introduction}

\label{sec:intro}
Deep learning has become state of the art solution for image classification problems having impressive accuracy but it is heavily impacted by objects characteristics like symmetry and rotations.

Most of the state of the art Convolutional Neural Networks (CNNs) were designed for training and classification using upright orientation \cite{AlexNet2012, He}. Their accuracy is heavily reduced if the objects in the image are rotated. While some data is naturally upright oriented (faces \cite{Huang} or numbers\cite{lecun-mnist2010}) other presents random orientations (plankton \cite{OrensteinBPS15}, galaxies \cite{GZoo}, food \cite{FooDB}).

To tackle the rotation problem, the majority of published approaches integrates rotated samples in the preparation of the training database. It is the case of MNIST-rot \cite{Larochelle2007} that contains random rotations on each sample of the training and validation set. Only some works, like ORNs \cite{Zhou2017} and the Rotationally-Invariant Convolution Module \cite{Follmann2018}, include results from networks trained with upright samples (MNIST) and validated on randomly rotated samples (MNIST-rot).
\begin{figure}[htbp]
	\centerline{\includegraphics[width=\linewidth]{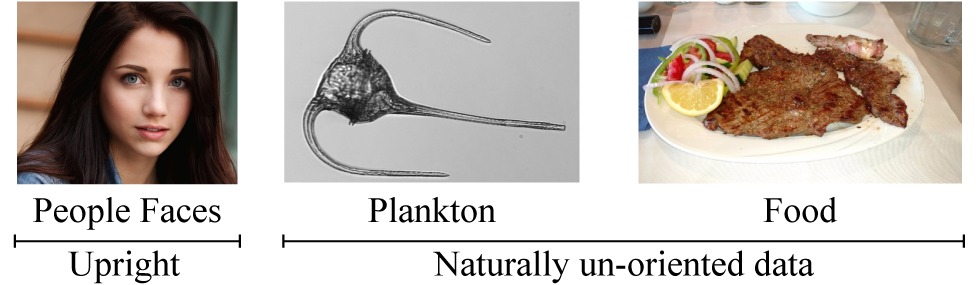}}
	\caption{Examples of upright and un-oriented datasets.}
	\label{samples} 
\end{figure}

In addition, the rotation problem has particular importance for some applications like automatic handling of industrial components by robotic arms require angular covariance \cite{Shin2006}. Here, the main challenge is not only classifying the rotated object but also to predict the angle without angle-annotated datasets.

In this paper, we present a rotation invariant architecture based on a Scattering Transform \cite{Bruna2013}. Having oriented wavelet feature space, we make use of the roto-translation properties it presents. Our architecture predicts the maximum probability class and its rotation. Furthermore, the network is able to continuously map the random rotation of the input to an output circular space. This circular space outputs the predicted angle despite being trained only with upright samples.

Our main contribution consists of an invariant and covariant CNN capable of predicting the class and the angle without angular labeling. We validate this feature by training the network with upright and randomly oriented samples achieving near the state of the art accuracy while reducing drastically the number of parameters.

This paper organization is: Section 2 discusses existing approaches and related work, in Section 3 we present the filter stage based on the Scattering Transform and discuss its roto-translational properties. Section 4 introduces the description of the convolutional architecture predictor, and finally we present the results of experiments in Section 5.

\section{Related work}
In the existing approaches that try to tackle the rotation problem, we can find two groups: i) transforming the input and ii) rotating the internal filters.

From the first group, data augmentation \cite{Dyk2001} is the most used method. It consists on generating random transformations of the input image, including change of size and rotations but it comes with some limitations. The number of filters increased when having sparse data to be able to capture the main features correctly. Also, the model still needs to learn a different filter set for each variation in the data. For example, different filters to detect horizontal and vertical edges.

We can cite other approaches included in this first group. The Spatial Transformer Network \cite{Jaderberg2015} and its variants, which apply a spatial transformation to feature maps. TI-Pooling \cite{Laptev2016} makes use of rotated versions of the same image as input and the network chooses the right rotation. Multi-Column Deep Neural Networks \cite{Ciregan2012}, train a model for each transformation and obtain the results by averaging and taking the winner-take-all output. Polar Transformer Networks \cite{Esteves2017} achieve rotation invariance by transforming the input into polar coordinates with the origin learned as the centroid of a single channel. The main limitation of these methods is to find a balance between the size of the network and the variations of the data.

The general objective of the approaches in the second group is to achieve rotation invariance by mean of internal filter rotations. They try to find a trade-off between required computational resources and the number of trainable parameters. Harmonic Networks \cite{Worrall2017} achieve equivariance to rotation by using steerable filters constructing any angular filter by the linear combination of base filters. Oriented Response Networks \cite{Zhou2017} propose the Active Rotating Filters that actively rotate during convolution and produce maps with location and orientation explicitly encoded. Rotation Equivariant Vector Field Networks \cite{Marcos2017} perform convolutions with several rotated instances of the same canonical filter but they rely on test-time data augmentation to improve their results. Recently, the Rotationally-Invariant Convolution Module \cite{Follmann2018} achieves rotation invariance by generating features that are rotationally invariant.

All of these approaches reach a good accuracy in terms of equivariance or invariance to the input and try to enhance the feature part of the network. In comparison to them, we present a new architecture that enhances the predictor part, allowing not only to predict class but also rotation. Also, we demonstrate the capability of the network to continuously map random rotations from training to a circular space containing the angle prediction.
\vspace{-0.3cm}
\section{Roto-translational feature space}
\begin{figure}[t]
	\centerline{\includegraphics[width=\linewidth]{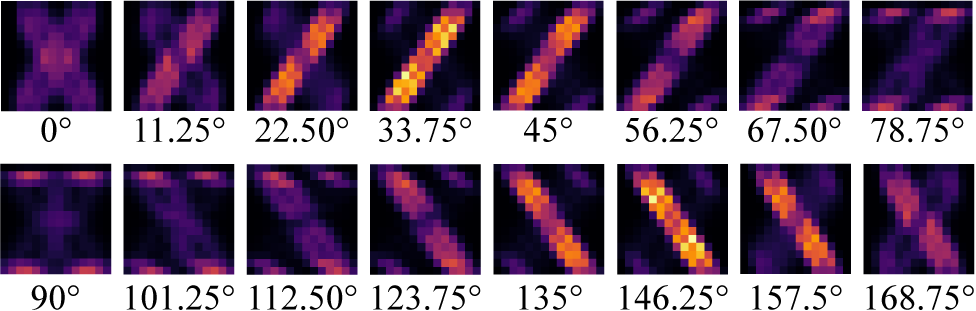}}
	\caption{Oriented-wavelet feature space for an upright input of the letter X product of the scattering transform (parameters $M = 2, J = 1, L = 16$).}
	\label{scatspace} 
\end{figure}

A wavelet scattering network computes a translation covariant image representation which is stable to deformations and preserves high-frequency information for classification. This network can provide the first layer of deep convolution networks \cite{Bruna2013}. We use the Scattering Transform to get an oriented-wavelet feature space (Fig. \ref{scatspace}). Translation in this space is covariant to rotation of the input (Fig. \ref{convarch}a).

\vspace{-0.5cm}
\subsection{Scattering wavelets}
\begin{figure*}[t]
	\centerline{\includegraphics[width=\linewidth]{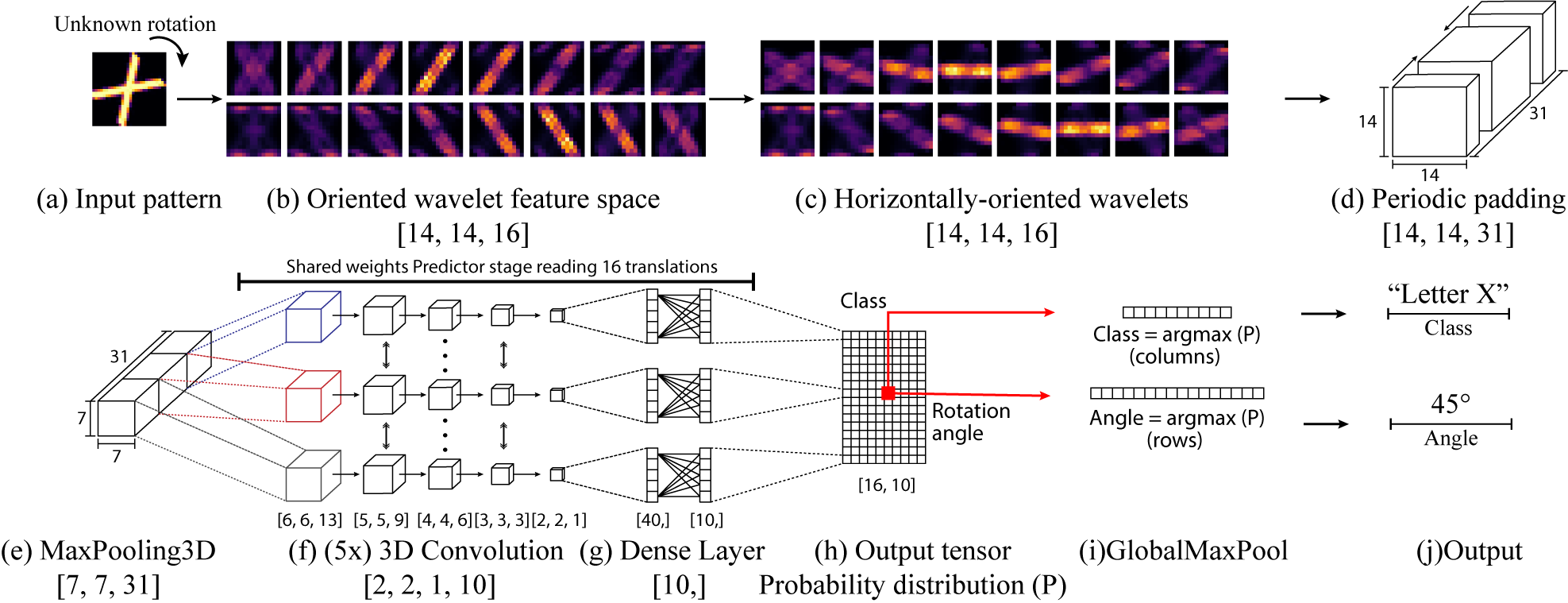}}
	\vspace{-0.3cm}
	\caption{Network architecture stages. First row, roto-translation features. Second row, overall predictor architecture. Output shape of each stage in brackets.}
	\vspace{-0.5cm}
	\label{convarch}
\end{figure*}

Using the real part of the Morlet wavelet described on \cite{Bruna2013} we transform the input into an oriented-wavelet feature space (Fig. \ref{convarch}b). This transformation outputs a series of wavelet samples with different energy each one. Angles that are colinear with edges in the input contain higher energy. For example, the letter X feature space will contain wavelets with more energy on the angles 33\textdegree\ and 146\textdegree\ and almost no energy at 0\textdegree. Furthermore, the distance between the angular strokes in the oriented wavelet feature space is proportional to the angular distance between the input edges.
\subsection{Roto-translation}
An important property of this space is the covariance between the rotation of the input and translation over the feature space. This translation is proportional to the angle $\theta$. Angular step $d\theta$ can be calculated by dividing the number of wavelet orientations ($n_s = 16$) present on the feature space by the angular range of the transform (180\textdegree). Another property of this transform is the capability of mapping the angular distance between the two image edges to a linear distance between the angular samples. Let the sample be the letter X containing an angle of 112\textdegree\ and other of 68\textdegree\ between the strokes. We can observe a linear distance of 10 steps between the angular samples for the bigger angle and of 6 steps for the smaller one. This linear distance remains constant for every rotation of the input image. The angular distance can be recovered by multiplying these numbers by $d\theta = 11.25$.

\subsection{Horizontally aligned wavelets}
While having the oriented wavelet space is important, the scan order of the image represents a crucial factor to achieve the rotation invariance and covariance over it, the wavelet angular sample should be scanned in the same orientation that it represents. 

To achieve this we implemented a custom weight dense layer which makes a re-indexing process over the feature space samples. This dense layer makes a bi-linear un-rotation that compensates the angle present in the wavelet using the oriented wavelet value calculated in the previous steps. 

The output result of this custom layer is an horizontally-oriented wavelet feature space (Fig. \ref{convarch}c) that contains the un-rotated version of each angular sample and that matches correctly with the horizontal scan order.
\vspace{-0.2cm}
\section{Convolutional predictor architecture}
\vspace{-0.2cm}
As a result of the roto-translation covariance, the wavelet feature space contains all the possible rotations of the input in the form of translations. To obtain all the translations we first apply periodic padding to the oriented wavelet feature space. The result of this periodic padding (Fig. \ref{convarch}d) is an augmented wavelet feature space with shape (14, 14, 31) containing all the possible translations.

To enhance the horizontally aligned wavelets information we apply a MaxPooling layer with size (2, 2, 1) (Fig. \ref{convarch}e). This reduces the number of parameters needed in the next layers. The output of this step is a tensor with shape (7, 7,~31).

To obtain information of each translation the predictor needs to span over the augmented wavelet feature space. This is, apply the predictor to the first 16 wavelet orientations and then move one step ahead. By applying the predictor in this way we obtain 16 different wavelet feature spaces of shape (7, 7, 16). Each one of these spaces contains one translation of the feature space.

The first stage of the predictor consists of five 3D Convolutions with kernel size (2, 2, 4) and 10 filters each one (Fig.~\ref{convarch}f). These convolutions capture the underlying features between the wavelet orientations. One of these features being the distance between them. This predictor is applied to each one of the spaces containing the translations. The importance of being a shared weight predictor spanning over them is to learn the features of the translation corresponding to the upright position. This upright position translation can appear on any of the translation spaces. The output of this stage are 16 spaces of (2, 2, 1, 10) containing the information of each translation.

The second stage of the predictor is a shared dense layer (Fig. \ref{convarch}g). This shared dense layer will be applied to each one of the output spaces from the first stage. The output shape of this dense layer is equal to the number of classes. This layer will make a prediction for each one of the translations and store it in a tensor. The output will be a probability distribution $P$ with as many columns as classes and rows as translations.

Let the case be $n_s = 16$ and 10 classes. The shared weight dense layer makes a prediction for each one of the 16 translations. Each one of these predictions is stored on the output tensor. This tensor has a shape of (16 x 10). The 10 columns of this tensor will contain the predicted class information and the 16 rows the angular information (Fig. \ref{convarch}h). After this, a GlobalMaxPooling layer (Fig. \ref{convarch}i) applied to the columns outputs the maximum probability class and its row index. Multiplying the row index by $d\theta$ plus a constant results in the predicted angle. 

\vspace{-0.7cm}
\section{Experiments}
\vspace{-0.2cm}
Following existing implementations of the state of the art, we validate the proposed architecture using the MNIST dataset. We made experiments with upright oriented samples and then with randomly oriented samples. Both of them validated on randomly rotated samples. To generate this variation of MNIST dataset we implemented a random rotation between [-90 and 90] for each sample of the original MNIST taking inspiration from MNIST-R \cite{Jaderberg2015}.
\vspace{-0.3cm}
\subsection{Scattering transform parameters}
We use a second order scattering transform $ M = 2 $, as suggested in \cite{Bruna2013}; higher order transforms are not useful because they have negligible energy. 

The scale parameter was fixed on $ J = 1$ as we are working with 28 x 28 pixels size images and the factor $ 2 ^ J$ makes the output image 14 x 14 pixels, further scaling of this parameter will reduce significantly the information available to the network. The last parameter was fixed at $L = 16$ allowing us to have 16 angular samples over -90\textdegree\ to 90\textdegree\ degrees range.
\vspace{-0.3cm}
\subsection{Rotation invariant class prediction}
Rotation invariance is validated by the capability of the architecture to predict the class correctly despite the rotation of the input. We test this property by training the network with MNIST-R (Table \ref{rrres}) and original MNIST (Table \ref{upres}). Both tests are validated on the randomly rotated dataset MNIST-R.  
\vspace{-0.3cm}
\begin{table}[htbp]
	\caption{Obtained error rate (training/validation=MNIST-R)}
	\begin{center}
		\begin{tabular}{|c|c|}
			\hline
			\textbf{Method}&\textbf{Error rate} \\
			\hline
			SVM \cite{Larochelle2007}& 10.38\%\\
			Harmonic Networks \cite{Worrall2017}& 1.69\% \\
			TI-Pooling \cite{Laptev2016}& 1.2\% \\
			Rotation Eq. Vector field networks\cite{Marcos2017} & 1.09\% \\
			ORN \cite{Zhou2017}& 0.76\% \\
			RP\_RF\_1* \cite{Follmann2018} & 3.51\% \\
			\hline
			Covariant CNN(Ours) & 2.69\% \\
			\hline
		\end{tabular}
		\label{rrres}
	\end{center}
\end{table}
\vspace{-0.7cm}
\begin{table}[htbp]
	\caption{Obtained error rate (training=MNIST; validation=MNIST-R)}
	\begin{center}
		\begin{tabular}{|c|c|}
			\hline
			\textbf{Method}&\textbf{Error rate} \\
			\hline
			ORN-8(ORPooling)\cite{Zhou2017}& 16.67\%\\
			ORN-8(ORAlign)\cite{Zhou2017}& 16.24\% \\
			\hline
			RotInv Conv. (RP\_RF\_1) \cite{Follmann2018} & 19.85\% \\
			RotInv Conv. (RP\_RF\_1\_32)* \cite{Follmann2018} & 12.20\% \\
			\hline
			Covariant CNN (Ours) & 17.21\% \\
			\hline
		\end{tabular}
		\label{upres}
	\end{center}
\end{table}

We can observe in Table \ref{rrres} that some accuracy is lost over the invariance as a cost of preserving the covariance over the network. However, this value is still approximate to state of the art implementations that are bigger in terms of trainable parameters and time to train. 

Table \ref{upres} demonstrates we have reached state of the art values in error rate with our method while using only 7,022 trainable parameters on the predictor stage. Is also worth to mention that RP\_RF\_1 has 130,050 parameters and RP\_RF\_1\_32 contains over 1 Million of trainable parameters.

\subsection{Prediction of the angle}
\label{sec:rotcov}
Due to variations in the data the predictor is tolerant to  slight variations of $\theta$. Consequently, it will output the maximum probability at the row corresponding to $\theta$ and a non-zero probability before and after (corresponding to $\theta\!\pm\!\delta\theta$).
This allows the predictor to map continuously the random rotation of the input to a circular space of the predicted angle.

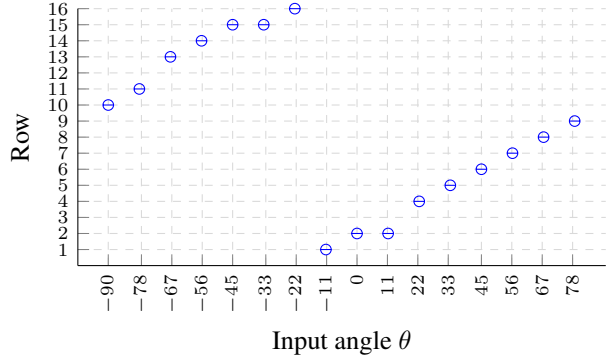
\begin{figure}[t]
	\begin{tikzpicture}
\begin{axis}[
enlarge y limits=false,
enlarge x limits=false,
axis x line*=middle,
axis y line*=left,
xmin=-101, xmax=90,
ymin=0, ymax=16,
xtick={-101, -90,-78,-67,-56,-45,-33,-22,-11,0,11,22,33,45,56,67,78},
ytick={1,...,15,16},
xlabel={Input angle $\theta$},
ylabel={Row},
grid = major,
grid style={dashed, gray!30},
width=\linewidth, height=5cm,    
yticklabel style = {font=\scriptsize},
xticklabel style = {font=\scriptsize, , rotate=90,anchor=east},
]
\centering
\addplot[blue, mark=halfcircle, only marks] table [y index = {1}, col sep=comma] {3Drr_1.txt};
\end{axis}
\end{tikzpicture}
\vspace{-0.4cm}
	\caption{The predicted rotation angle is mapped into a continuous, circular space.}
	\label{fig:linearcov} 
	\vspace{-0.5cm}
\end{figure}

We test this by rotating an input sample in steps of $d\theta = 11.25$\textdegree\ from [-90\textdegree\ to 78.75\textdegree] and plotting the output circular space represented by the rows of the output prediction (Fig.\ref{fig:linearcov}). The rotation is mapped to the rows of the output tensor described previously. The output exhibits a self-organizing behavior of mapping consecutive angular values as consecutive rows in the table. This comes as a result of non-zero class probability on $\theta$$\pm$$\delta\theta$ with maximum probability on $\theta$ and lower on the previous and next angular steps. When the absolute angular reference is unknown (e.g. for plankton upright position does not exist) the network maps one of the rotation values to one point of the linear space and then the consecutive angles are linearly mapped.

The linear output space contains the angular information of the input. This space has the same properties and behavior when trained with upright oriented datasets and randomly rotated datasets. This leads to generating a linear relationship from the consecutive angles without any reference existing on the angular rotation input space.

\vspace{-0.4cm}
\section{Conclusions}
\vspace{-0.2cm}
We demonstrate the capability of obtaining rotation invariance by training the network with only upright samples. The network is capable of predicting angles unseen on the training phase. In addition, when the input data is naturally random oriented the architecture is able to infer the orientation of the samples and generate a linear relationship between them. This allows further applications of this architecture for automatic alignment of randomly oriented datasets.

We reached near state of the art error rate values while using one feature calculated by the scattering transform. We expect to have a relatively small size in the network for input images bigger than the presented on this work (28 x 28), tests in bigger images have shown increased inference time caused by the scanning and prediction of every translation contained in the feature space.

The next step to improve the error rate is to replace the scattering transform by a trainable feature stage that preserves this  roto-translational property and validate it on other datasets like plankton, food or faces.
\bibliographystyle{IEEEbib}
\bibliography{strings,refs}

\end{document}